\documentclass{article}

\usepackage{arxiv}

\usepackage[utf8]{inputenc} 
\usepackage[T1]{fontenc}    
\usepackage{hyperref}       
\usepackage{url}            
\usepackage{booktabs}       
\usepackage{amsfonts}       
\usepackage{nicefrac}       
\usepackage{microtype}      
\usepackage{lipsum}
\usepackage{graphicx}
\graphicspath{ {./images/} }

\usepackage{amssymb}
\usepackage{amsmath}
\usepackage{subcaption}

\title{DualFlexKAN: Dual-stage Kolmogorov-Arnold Networks with Independent Function Control}

\author{
 Andrés Ortiz \\
  Department of Communications Engineering\\
  University of Malaga\\
  Malaga, Spain \\
     DaSCI Andalusian Institute of Data Science \\
   and Computational Intelligence\\
  University of Granada\\
  Granada, Spain \\
  \texttt{andres.ortiz@uma.es} \\
   \And
 Nicolás J. Gallego-Molina \\
  Department of Communications Engineering\\
  University of Malaga\\
  Malaga, Spain \\
     DaSCI Andalusian Institute of Data Science \\
   and Computational Intelligence\\
  University of Granada\\
  Granada, Spain \\
  \And
 Carmen Jiménez-Mesa \\
  Department of Communications Engineering\\
  University of Malaga\\
  Malaga, Spain \\
   \And
 Juan M. Górriz\\
  Department of signal Theory, Communications and Telematics\\
  University of Granada\\
  Granada, Spain \\
     DaSCI Andalusian Institute of Data Science \\
   and Computational Intelligence\\
  University of Granada\\
  Granada, Spain \\
     \And
 Javier Ramírez \\
  Department of signal Theory, Communications and Telematics\\
  University of Granada\\
  Granada, Spain \\
   DaSCI Andalusian Institute of Data Science \\
   and Computational Intelligence\\
  University of Granada\\
  Granada, Spain \\
}

\begin{document}
\maketitle
\begin{abstract}
Multi-Layer Perceptrons (MLPs) rely on pre-defined, fixed activation functions, 
imposing a static inductive bias that forces the network to approximate complex 
topologies solely through increased depth and width. Kolmogorov-Arnold Networks 
(KANs) address this limitation through edge-centric learnable functions, yet their 
formulation suffers from quadratic parameter scaling and architectural rigidity that 
hinders the effective integration of standard regularization techniques. This paper 
introduces the DualFlexKAN (DFKAN), a flexible architecture featuring a dual-stage mechanism 
that independently controls pre-linear input transformations and post-linear output 
activations. This decoupling enables hybrid networks that optimize the trade-off 
between expressiveness and computational cost. Unlike standard formulations, 
DFKAN supports diverse basis function families, including orthogonal 
polynomials, B-splines, and radial basis functions, integrated with configurable 
regularization strategies that stabilize training dynamics. Comprehensive evaluations 
across regression benchmarks, physics-informed tasks, and function approximation 
demonstrate that DFKAN outperforms both MLPs and conventional KANs in 
accuracy, convergence speed, and gradient fidelity. The proposed hybrid 
configurations achieve superior performance with one to two orders of magnitude 
fewer parameters than standard KANs, effectively mitigating the parameter explosion 
problem while preserving KAN-style expressiveness. DFKAN provides a 
principled, scalable framework for incorporating adaptive non-linearities, proving 
particularly advantageous for data-efficient learning and interpretable function 
discovery in scientific applications.
\end{abstract}


\section{Introduction}
\label{sec:intro}
The emergence of neural networks derived from Multi-layer Perceptrons (MLPs) and their expansion to deep networks in the 2000s revolutionised classification and pattern recognition applications, significantly enhancing the performance of models based on statistical learning algorithms such as Support Vector Machines (SVMs) \cite{cortes1995SVM}. Neural-based models excel particularly in cases where input–output relationships are highly nonlinear, high-dimensional, or subject to complex interaction structures. In contrast, SVMs rely on predefined kernel functions that often struggle with large datasets, whereas MLPs use data-driven feature representations that automatically capture intricate interactions without manual feature engineering. For instance, deep frameworks have outperformed linear and kernel-based methods, including SVM variants, in multi-trait genome-wide prediction by modelling non-additive and epistatic effects that classical approaches struggle to represent \cite{FanWaldmann2024}. Similarly, well-regularised MLPs have surpassed both classical Machine Learning (ML) models and more elaborate neural architectures on tabular benchmarks, revealing that deep networks learn richer latent structures than SVMs, which depend heavily on kernel choice and hyperparameter sensitivity \cite{KadraLindauerHutterGrabocka2021}. Improvements have also been made using hybrid deep architectures that combine tree-based inductive biases with MLPs, resulting in better performance compared to strong baselines that include SVMs and gradient-boosting methods across diverse tabular tasks \cite{YanChenWangChenWu2024}. Comprehensive benchmarking confirms the broad advantage of deep networks in heterogeneous tabular settings when appropriately tuned \cite{ShmuelGlickmanLazebnik2024}. Additionally, deep MLP-like models have shown superior performance in biomedical applications, such as disease prediction from metagenomic data, where highly nonlinear feature interactions render SVMs comparatively less expressive \cite{JiangYangPeng2024}.

The \textit{magic} of MLP-based networks resides in their layered structure of neurons that apply learned weights and nonlinear activations. Each layer transforms inputs into progressively abstract representations, enabling the network to capture intricate feature interactions and hierarchical patterns, thus modelling complex non-linear relationships that linear models or fixed-kernel methods cannot effectively represent \cite{cybenko1989approximation,hornik1989multilayer}. This model has been successful across a wide range of domains, including deep learning (DL). It restricts the expressiveness of neural networks by only allowing nonlinearity in predetermined forms. Recent theoretical developments in neural network approximation have motivated a fundamental reconsideration of this architectural paradigm.

Kolmogorov-Arnold Networks (KANs) represent a paradigm shift in neural network design, which draws inspiration from the Kolmogorov-Arnold representation theorem~\cite{kolmogorov1957representation,arnold1957functions,sprecher1965kolmogorov}. Unlike conventional multilayer perceptrons, which apply uniformly fixed activation functions across their layers, knowledge-aware networks use learnable one-variable functions on the edges of the computational graph, enabling the network to adaptively identify optimal nonlinear transformations during training~\cite{liu2024kan}. This innovation fundamentally changes the learning focus: whereas MLPs learn through weight matrices with fixed activation values, KAN learn the activation functions themselves, parameterised through flexible approximators such as splines, polynomial expansions, or orthogonal basis functions. 

The theoretical basis of KANs offers several advantages over conventional MLP architectures. KANs can achieve greater expressiveness with fewer parameters by enabling the network to learn suitable nonlinear relationships rather than imposing them beforehand, which may lead to more streamlined and efficient models \cite{liu2024kan}. The explicit representation of learned functions improves interpretability, as it allows for the examination of individual univariate transformations to analyse the network's behaviour. Practitioners can select from various basis functions, including B-splines and orthogonal polynomial basis, to incorporate problem-specific inductive biases or relevant mathematical structures.

Despite the theoretical advantages of KAN, practical implementations have been hindered by significant challenges, restricting their widespread adoption~\cite{hou2024critical}. Employing per-edge or per-neuron function parameterizations leads to significantly increased parameter spaces, primarily as a result of increased flexibility in function representation. The increase in parameters not only amplifies computational requirements but also sparks concerns about stability and overfitting, especially in cases where data is scarce. The selection of basis functions and their parameterization introduces extra hyperparameters that need to be carefully adjusted, thereby complicating the model selection process. The training dynamics of KANs can exhibit instability due to the simultaneous optimisation of both linear weights and function parameters, necessitating careful initialisation strategies~\cite{glorot2010understanding,he2015delving} and potentially customised optimisation methods. 

A significant drawback of current KAN implementations is their inflexibility in terms of architecture. The initial formulations of these strategies usually enforce a consistent function-sharing approach across the network. However, they often overlook the fact that different layers or network positions may require different levels of adaptability. For example, early layers processing raw input features may require different transformation properties than deeper layers operating on higher-level representations~\cite{he2015deep}. Replacing linear transformations entirely with learnable functions may be overly aggressive in scenarios where linear operations are sufficient or where computational efficiency must take precedence.

These observations motivated the development of DualFlexKAN, a flexible framework that bridges the gap between traditional MLPs and full KANs. DualFlexKAN introduces a dual-stage architecture with independent control over input transformation functions and output activation functions, coupled with configurable function-sharing strategies at both stages. This design philosophy allows practitioners to build hybrid architectures by applying learnable functions where they are most beneficial, ensuring computational efficiency through fixed or shared transformations at other points. The framework accommodates a variety of basis function families, including standard polynomials, Legendre polynomials, Gegenbauer polynomials, Jacobi polynomials, B-splines, and radial basis functions, each of which provides distinct approximation properties and inductive biases.

The key innovation of DualFlexKAN is its granular control over function adaptability. The input stage of the framework supports five distinct strategies: no transformation to maintain raw inputs, fixed non-learnable functions for prescribed preprocessing, globally shared learnable functions to capture common input patterns, per-input functions for feature-specific transformations, and per-neuron-input functions that enable maximum expressiveness. Similarly, the output stage provides four strategies ranging from no activation to per-neuron learnable activations. This hierarchical configuration space encompasses both traditional MLPs and full KANs as specific instances, facilitating a diverse range of intermediate network architectures. 

Moreover, DualFlexKAN incorporates flexible regularisation mechanisms designed to address the stability and overfitting challenges inherent in high-capacity function learning. This framework allows practitioners to position dropout~\cite{srivastava2014dropout} and batch normalisation~\cite{ioffe2015batch} at strategic points within the computational graph, either immediately before, immediately after, or both with respect to activations, depending on a specific sequence when multiple regularisation techniques are employed simultaneously. This fine-grained control enables the examination of customised regularisation methods designed to fit the distinct training characteristics of learnable activation functions. 

This paper presents a comprehensive study of the DualFlexKAN architecture, examining its theoretical properties, computational characteristics, and empirical performance across diverse learning tasks. We develop tailored initialization methods that take into account the interconnected dynamics of linear and functional parameters, suggest criteria for choosing suitable basis functions and data sharing approaches based on problem specifics, and offer systematic comparisons against both standard MLPs and existing KAN implementations. We show through comprehensive experimentation that precisely designed DualFlexKAN architectures can realize improved performance-efficiency trade-offs, providing professionals with a structured method for integrating trainable non-linearities into neural networks while upholding practical manageability.

The remainder of this paper is organized as follows. Section~\ref{sec:methods} reviews related work on learnable activation functions and the Kolmogorov-Arnold representation theorem. Section~\ref{sec:mathematical_KAN} presents the mathematical formulation of DualFlexKAN, its architectural variants, and our initialization and training methodologies. Section~\ref{sec:results} reports experimental results on benchmark datasets, along with a detailed analysis and discussion. Finally, Section~\ref{sec:conclusions} concludes the paper and outlines directions for future research.

\section{Multi-Layer Perceptrons versus Kolmogorov-Arnold Networks}
\label{sec:methods}

\subsection{Multi-Layer Perceptron Architecture}

\begin{figure}[h!]
    \centering
    \includegraphics[width=0.9\textwidth]{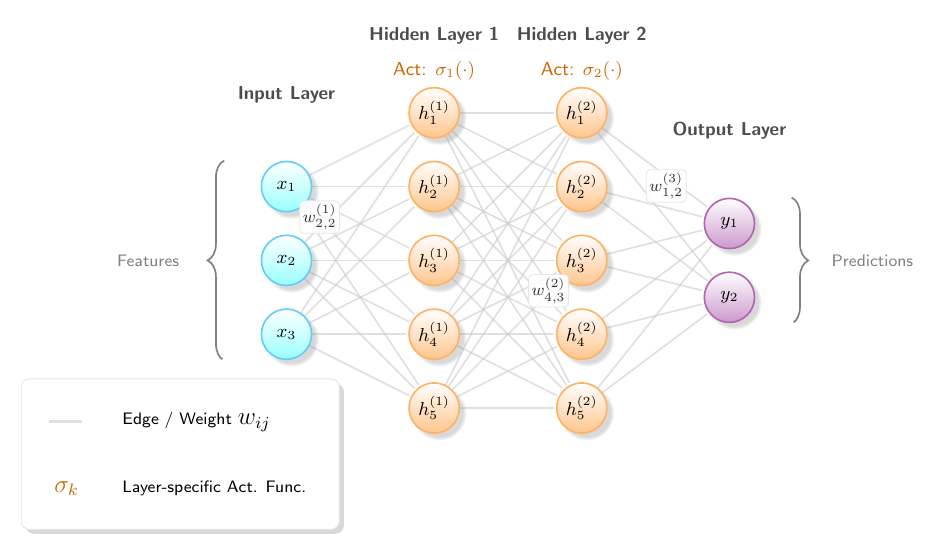}
    \caption{Classical MLP architecture}
    \label{fig:MLP_figure}
\end{figure}

Multi-Layer Perceptrons constitute the fundamental building blocks of deep learning architectures~\cite{goodfellow2016deep,lecun2015deep}. An MLP with $L$ layers implements a composition of affine transformations followed by fixed non-linear activation functions. The basic architecture of an MLP with two hidden layers is shown in  Figure \ref{fig:MLP_figure}. For an input vector $\mathbf{x} \in \mathbb{R}^{n_0}$ representing a sample in the feature space, the forward propagation through layer $l$ is defined as:

\begin{equation}
\mathbf{z}^{(l)} = \sigma^{(l)}\left(\mathbf{W}^{(l)} \mathbf{z}^{(l-1)} + \mathbf{b}^{(l)}\right), \quad l = 1, 2, \ldots, L
\end{equation}

\noindent where $\mathbf{W}^{(l)} \in \mathbb{R}^{n_l \times n_{l-1}}$ denotes the weight matrix, $\mathbf{b}^{(l)} \in \mathbb{R}^{n_l}$ represents the bias vector, $\mathbf{z}^{(0)} = \mathbf{x}$, and $\sigma^{(l)}: \mathbb{R} \to \mathbb{R}$ is a fixed activation function applied element-wise. The output of the network is $\mathbf{y} = \mathbf{z}^{(L)}$.

Common choices for $\sigma$ include the Rectified Linear Unit (ReLU)~\cite{nair2010rectified}, hyperbolic tangent, sigmoid, and their variants~\cite{ramachandran2017searching}. The Universal Approximation Theorem (UAT) \cite{cybenko1989approximation,hornik1989multilayer} establishes that MLPs with a single hidden layer of sufficient width can approximate any continuous function on compact subsets of $\mathbb{R}^n$ to arbitrary precision. Formally, for any continuous function $f: [0,1]^n \to \mathbb{R}$ and $\epsilon > 0$, there exists an MLP such that:

\begin{equation}
\sup_{\mathbf{x} \in [0,1]^n} |f(\mathbf{x}) - \text{MLP}(\mathbf{x})| < \epsilon
\end{equation}

\subsubsection*{Relation to representation theory}
It is critical to distinguish the Universal \textit{Approximation} Theorem utilized by MLPs from the Universal \textit{Representation} Theorem. While the Kolmogorov-Arnold representation guarantees that a multivariate function can be written \textit{exactly} using a finite number of variable univariate functions, the MLP UAT achieves this via an \textit{approximation} using an arbitrarily large number of fixed basis functions. Theoretically, an MLP seeks to satisfy the representation theorem by constructing the necessary univariate inner and outer functions ($\phi$ and $\Phi$) through linear combinations of its fixed non-linearities (e.g., ReLUs). Consequently, to approximate the complex, often non-smooth univariate functions required by the representation theorem, the MLP is forced to increase its width $n_l \to \infty$. This reliance on width to compensate for fixed activation bases is a primary source of inefficiency in standard MLPs.

Despite this theoretical expressiveness, MLPs face practical limitations. The fixed nature of activation functions constrains the network's ability to adaptively learn optimal non-linear transformations specific to the task at hand. All neurons in a given layer share the same activation function type, preventing heterogeneous function learning that might better capture data structure. Furthermore, the number of parameters scales as $\mathcal{O}(n_l \cdot n_{l-1})$ per layer, potentially leading to overparameterization in deep architectures.

\subsection{Kolmogorov-Arnold Networks: Learnable Activation Functions}

\begin{figure}[h!]
    \centering
    \includegraphics[width=0.8\textwidth]{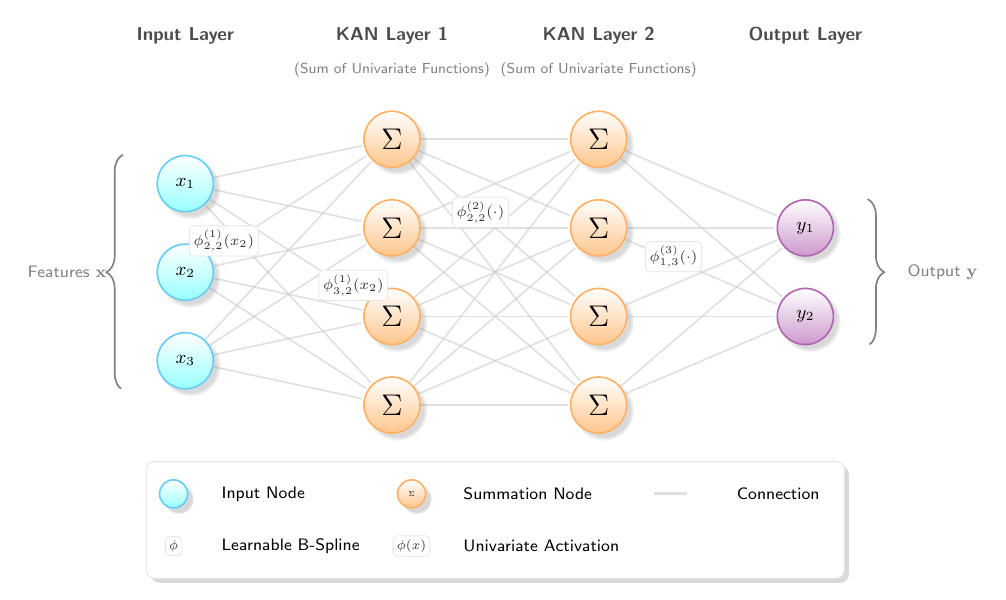}
    \caption{Classical KAN architecture}
    \label{fig:KAN_figure}
\end{figure}

The Kolmogorov-Arnold representation theorem~\cite{kolmogorov1957representation,arnold1957functions} provides an alternative theoretical foundation for neural network design. The theorem states that any multivariate continuous function $f: [0,1]^n \to \mathbb{R}$ can be represented as:

\begin{equation}
f(\mathbf{x}) = f(x_1, x_2, \ldots, x_n) = \sum_{q=0}^{2n} \Phi_q \left( \sum_{p=1}^{n} \phi_{q,p}(x_p) \right)
\end{equation}

\noindent where $\phi_{q,p}: [0,1] \to \mathbb{R}$ are univariate inner functions and $\Phi_q: \mathbb{R} \to \mathbb{R}$ are univariate outer functions. This theorem demonstrates that high-dimensional functions can be constructed through compositions and summations of one-dimensional functions, a structure fundamentally different from the MLP paradigm. An example of a classical KAN architecture is shown in Figure \ref{fig:KAN_figure}, where the connections are modeled using b-spline.

Inspired by this representation, KANs~\cite{liu2024kan} replace the fixed activation functions in MLPs with learnable univariate functions on network edges. A KAN layer transforms an input $\mathbf{x} \in \mathbb{R}^{n_{\text{in}}}$ to output $\mathbf{y} \in \mathbb{R}^{n_{\text{out}}}$ via:

\begin{equation}
y_j = \sum_{i=1}^{n_{\text{in}}} \phi_{i,j}(x_i), \quad j = 1, 2, \ldots, n_{\text{out}}
\end{equation}

\noindent where each $\phi_{i,j}: \mathbb{R} \to \mathbb{R}$ is a learnable univariate function. Notably, KANs eliminate traditional weight matrices entirely, replacing linear combinations with learned function evaluations. The functions $\phi_{i,j}$ are typically parameterized using flexible basis expansions such as B-splines~\cite{liu2024kan}, enabling smooth, differentiable approximations learned via backpropagation.

For a B-spline parameterization of order $k$ with control points $\{c_i\}_{i=1}^{m}$ and knot sequence $\{t_i\}$, the learned function takes the form:

\begin{equation}
\phi(x) = \sum_{i=1}^{m} c_i B_{i,k}(x)
\end{equation}

\noindent where $B_{i,k}$ denotes the $i$-th B-spline basis function of order $k$. The control points $\{c_i\}$ constitute the learnable parameters optimized during training.

\subsection{Comparative Analysis and Motivation}

The fundamental distinction between MLPs and KANs lies in the locus of learning and their relationship to the curse of dimensionality. MLPs learn through weight matrices with predetermined activation functions, parametrizing the network as:

\begin{equation}
\text{MLP: } \mathbf{y} = \sigma(\mathbf{W}\mathbf{x} + \mathbf{b}) \quad \text{with learnable } \{\mathbf{W}, \mathbf{b}\}
\end{equation}

In contrast, KANs learn the activation functions themselves while eliminating traditional weights:

\begin{equation}
\text{KAN: } y_j = \sum_{i=1}^{n} \phi_{i,j}(x_i) \quad \text{with learnable } \{\phi_{i,j}\}
\end{equation}

This architectural shift offers several theoretical advantages. First, KANs align directly with the Kolmogorov-Arnold representation theorem, treating the univariate functions $\phi$ as the primary learnable entities rather than approximating them via hidden layers. This allows KANs to potentially escape the curse of dimensionality: while MLPs often require exponentially growing widths to approximate high-dimensional functions (codifying the representation theorem's inner functions), KANs can achieve comparable accuracy with a parameter count scaling as $\mathcal{O}(n^2)$ by increasing the complexity (grid size) of the spline bases rather than the network width~\cite{liu2024kan}. Second, the explicit representation of learned functions enhances interpretability, enabling direct visualization and analysis of the discovered non-linear transformations~\cite{xu2024kan}. Third, the flexibility in basis function selection, including wavelets~\cite{bozorgasl2024wavkan}, Chebyshev polynomials~\cite{shukla2024comprehensive}, and sinusoidal functions~\cite{aghazadeh2024sinekan} allows incorporation of domain-specific inductive biases. 

However, KANs introduce significant practical challenges~\cite{hou2024critical}. The parameter count for a KAN layer with $n_{\text{in}}$ inputs, $n_{\text{out}}$ outputs, and $m$ basis functions per edge scales as $\mathcal{O}(n_{\text{in}} \cdot n_{\text{out}} \cdot m)$, potentially exceeding MLP parameterization. Training stability can be compromised due to coupled optimization of numerous univariate functions. Moreover, existing implementations enforce uniform function-sharing strategies across all layers, limiting architectural heterogeneity that might better balance expressiveness and efficiency.

To address these fundamental limitations, we introduce the DualFlex Kolmogorov-Arnold Network (DFKAN). By decoupling the non-linear transformations into pre-linear (input) and post-linear (output) stages, DFKAN establishes a flexible hierarchical design space. Compared to conventional edge-centric KANs, DFKAN provides four distinct structural and operational advantages:

\begin{enumerate}
    \item \textbf{Avoiding Parameter Explosion (Structural Efficiency):} Conventional KANs require $\mathcal{O}(n_{\text{in}} \cdot n_{\text{out}} \cdot m)$ parameters per layer, rendering deep or wide architectures computationally prohibitive. By strategically applying shared (\textit{global}) or \textit{per-dimension} function strategies, DFKAN breaks this combinatorial scaling, achieving competitive expressiveness with a parameter footprint comparable to standard MLPs (often yielding 1 to 2 orders of magnitude fewer parameters than vanilla KANs).
    
    \item \textbf{Overcoming the \textit{Additive Bottleneck} via Efficient Depth:} Shallow KANs inherently struggle to capture multiplicative or highly nested interactions (e.g., $\sin(2x)\cos(2y)$) due to their additive separability bias. While deepening a vanilla KAN introduces severe training instability, DFKAN's node-centric efficiency allows for the seamless insertion of deeper interaction layers, accurately capturing continuous differentiable topologies and high-frequency gradients without destabilizing the network.
    
    \item \textbf{Inherent Regularization and Noise Robustness (\textit{Occam's Razor}):} The dense, per-edge parameterization of vanilla KANs makes them highly susceptible to overfitting spurious correlations and high-frequency noise in low-data regimes. By constraining plasticity through shared activation manifolds, DFKAN acts as a structural regularizer. It naturally filters stochastic noise to recover the underlying smooth physical laws, proving vastly superior for symbolic regression and scientific discovery tasks.
    
    \item \textbf{Architectural Heterogeneity and Biological Plausibility:} Unlike vanilla KANs, which force a uniform parameterization across the entire network, DFKAN enables layer-specific adaptability. Practitioners can design networks where early layers utilize highly expressive \textit{per-connection} functions (mimicking complex dendritic computations for feature extraction), while deeper layers transition to stable, shared functions (mimicking somatic integration for decision-making). Furthermore, DFKAN allows the seamless integration of standard regularization techniques (Dropout and Batch Normalization), which are challenging to implement effectively in edge-centric KANs.
\end{enumerate}

\section{DualFlexKAN: Dual-Stage Flexible Kolmogorov-Arnold Networks}
\label{sec:mathematical_KAN}

\subsection{Architecture Overview}

\begin{figure}[h!]
    \centering
    \includegraphics[width=0.9\textwidth]{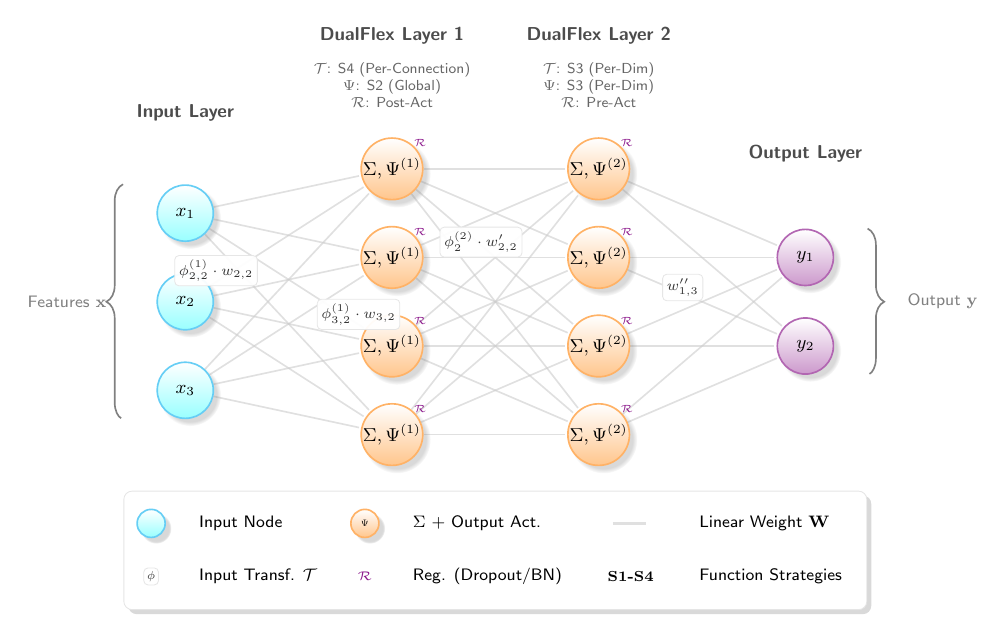}
    \caption{DualFlexKAN architecture. Some configuration options are shown in the Figure.}
    \label{fig:DFKAN_figure}
\end{figure}

DualFlexKAN introduces a dual-stage architecture that decouples input transformation functions from output activation functions, providing independent and granular control over each stage. Unlike standard KANs that apply learnable functions solely on network edges, DualFlexKAN separates the transformation process into two distinct phases: pre-linear input transformation and post-linear output activation. This separation enables hybrid architectures that strategically balance expressiveness, computational efficiency, and parameter count on a per-layer basis. A graphical representation of the DualFlexKAN architecture along with the most representative configuration options is depicted in Figure \ref{fig:DFKAN_figure}.

\subsubsection*{Neurobiological Motivation} The dual-stage decoupling introduced by DFKAN finds strong parallels in modern neurobiology, moving beyond the oversimplified point-neuron model used in classical MLPs. In biological neurons, synaptic inputs are not merely weighted and summed. They undergo complex, localized non-linear transformations along the dendritic tree before reaching the cell body (dendritic computation). This is mathematically mirrored by DFKAN's pre-linear input transformations, where employing a \textit{per-connection} strategy simulates the highly adaptable nature of dendritic branches. Conversely, the post-linear output activation mimics the somatic integration and the subsequent action potential generated at the axon hillock, a mechanism that is generally more stereotyped and acts as a global thresholding function. 

Furthermore, DFKAN's architectural flexibility to transition from highly expressive to constrained functions across layers mirrors the hierarchical processing in the human brain. Early layers configured with \textit{per-connection} or \textit{per-input} learnable functions analogous to the early sensory cortices (e.g., primary visual cortex), which require immense plasticity and diverse receptive fields to extract intricate patterns from raw stimuli. As information propagates deeper into the network, transitioning to \textit{global} or \textit{fixed} activation strategies reflects the behavior of higher-order association cortices. In these deeper regions, the brain integrates highly abstracted, pre-processed features using stable, generalized mechanisms to form robust cognitive representations and logical decisions. By mimicking this biological hierarchy, DFKAN achieves both high representational capacity in feature extraction and stable, generalizable integration in decision layers, simultaneously preventing the parameter explosion seen in vanilla KANs.

\subsection{Mathematical Formulation}

For a DualFlexKAN layer $l$ with $n_{\text{in}}$ input dimensions and $n_{\text{out}}$ output dimensions, the forward propagation is defined as:

\begin{equation}
\mathbf{z}^{(l)} = \mathcal{R}_{\text{out}}^{(l)} \left( \Psi^{(l)} \left( \mathbf{W}^{(l)} \mathcal{T}^{(l)}(\mathbf{z}^{(l-1)}) + \mathbf{b}^{(l)} \right) \right)
\end{equation}

\noindent where:
\begin{itemize}
\item $\mathcal{T}^{(l)}: \mathbb{R}^{n_{\text{in}}} \to \mathbb{R}^{n_{\text{in}}}$ is the input transformation operator
\item $\mathbf{W}^{(l)} \in \mathbb{R}^{n_{\text{out}} \times n_{\text{in}}}$ is the linear weight matrix
\item $\mathbf{b}^{(l)} \in \mathbb{R}^{n_{\text{out}}}$ is the bias vector
\item $\Psi^{(l)}: \mathbb{R}^{n_{\text{out}}} \to \mathbb{R}^{n_{\text{out}}}$ is the output activation operator
\item $\mathcal{R}_{\text{out}}^{(l)}$ is the post-activation regularization operator
\end{itemize}

The complete layer computation can be expanded as:

\begin{equation}
\mathbf{s}^{(l)} = \mathbf{W}^{(l)} \mathcal{T}^{(l)}(\mathbf{z}^{(l-1)}) + \mathbf{b}^{(l)}
\end{equation}

\begin{equation}
\mathbf{a}^{(l)} = \mathcal{R}_{\text{pre}}^{(l)}(\mathbf{s}^{(l)})
\end{equation}

\begin{equation}
\mathbf{z}^{(l)} = \mathcal{R}_{\text{post}}^{(l)} \left( \Psi^{(l)}(\mathbf{a}^{(l)}) \right)
\end{equation}

\noindent where $\mathcal{R}_{\text{pre}}^{(l)}$ and $\mathcal{R}_{\text{post}}^{(l)}$ denote pre-activation and post-activation regularization operators, respectively.

\subsection{Function Sharing Strategies}

Both input transformation $\mathcal{T}^{(l)}$ and output activation $\Psi^{(l)}$ operators can be configured independently using one of the following strategies. For notational clarity, we denote a generic function operator as $\mathcal{F}: \mathbb{R}^{n} \to \mathbb{R}^{n}$ and its component-wise evaluation as $[\mathcal{F}(\mathbf{x})]_i$ for dimension $i$.
\smallskip

\textbf{Strategy 0: None (Identity/Linear).} No transformation is applied:
\begin{equation}
\mathcal{F}(\mathbf{x}) = \mathbf{x}
\end{equation}
Parameter count: $N_{\mathcal{F}} = 0$.

\textbf{Strategy 1: Fixed Function.} A predetermined non-learnable function $f: \mathbb{R} \to \mathbb{R}$ is applied element-wise:
\begin{equation}
[\mathcal{F}(\mathbf{x})]_i = f(x_i), \quad i = 1, \ldots, n
\end{equation}
Common choices include $f(x) = \text{ReLU}(x)$, $f(x) = \tanh(x)$, or $f(x) = \sigma(x)$. Parameter count: $N_{\mathcal{F}} = 0$.

\smallskip

\textbf{Strategy 2: Global Shared Function.} A single learnable function $\phi_{\text{global}}: \mathbb{R} \to \mathbb{R}$ is shared across all dimensions:
\begin{equation}
[\mathcal{F}(\mathbf{x})]_i = \phi_{\text{global}}(x_i), \quad i = 1, \ldots, n
\end{equation}
Parameter count: $N_{\mathcal{F}} = m$, where $m$ is the number of basis function coefficients.
\smallskip

\textbf{Strategy 3: Per-Dimension Function.} Each dimension has its own learnable function $\phi_i: \mathbb{R} \to \mathbb{R}$:
\begin{equation}
[\mathcal{F}(\mathbf{x})]_i = \phi_i(x_i), \quad i = 1, \ldots, n
\end{equation}
Parameter count: $N_{\mathcal{F}} = n \cdot m$.
\smallskip

\textbf{Strategy 4: Per-Connection Function (Input Only).} For the input transformation operator $\mathcal{T}^{(l)}$, each connection from input $i$ to output neuron $j$ has a unique learnable function $\phi_{i,j}: \mathbb{R} \to \mathbb{R}$. The transformed input contributes to the linear combination as:
\begin{equation}
s_j^{(l)} = \sum_{i=1}^{n_{\text{in}}} w_{j,i}^{(l)} \phi_{i,j}(z_i^{(l-1)}) + b_j^{(l)}
\end{equation}
Parameter count: $N_{\mathcal{F}} = n_{\text{in}} \cdot n_{\text{out}} \cdot m$. This strategy provides maximum expressiveness and is applicable only to input transformations.

The choice of strategy for $\mathcal{T}^{(l)}$ and $\Psi^{(l)}$ can be configured independently and may differ across layers, enabling fine-grained architectural design. For instance, layer $l$ might employ strategy S2 (global) for input transformation and strategy S3 (per-dimension) for output activation, while layer $l+1$ uses strategy S1 (fixed) for both stages.

\subsection{Basis Function Families}

Learnable functions in DualFlexKAN are parameterized using flexible basis expansions. Each learnable function $\phi: \mathbb{R} \to \mathbb{R}$ is expressed as:

\begin{equation}
\phi(x) = \sum_{k=1}^{m} c_k \mathcal{B}_k(x; \boldsymbol{\theta})
\end{equation}

\noindent where $\{c_k\}_{k=1}^{m}$ are learnable coefficients, $\mathcal{B}_k$ are basis functions, and $\boldsymbol{\theta}$ represents optional basis-specific parameters.

\begin{table}[htbp]
\centering
\caption{Basis Function Families in DualFlexKAN}
\label{tab:basis_functions}
\small
\begin{tabular}{p{2.5cm}p{3cm}p{1.8cm}p{2cm}p{3.5cm}}
\toprule
\textbf{Basis Type} & \textbf{Definition} & \textbf{Domain} & \textbf{Weight $w(x)$} & \textbf{Learnable Parameters} \\
\midrule
\textbf{Standard} \textbf{Polynomial} & $\mathcal{B}_k(x) = x^{k-1}$ & $\mathbb{R}$ & N/A & $\{c_k\}_{k=1}^m$ \\
\addlinespace
\textbf{Legendre} & $P_n(x)$ via recurrence: $P_0=1$, $P_1=x$ & $[-1, 1]$ & $w(x) = 1$ & $\{c_k\}_{k=1}^m$ \\
\addlinespace
\textbf{Chebyshev} & $T_n(x)$ via recurrence: $T_0=1$, $T_1=x$ & $[-1, 1]$ & $w(x) = \frac{1}{\sqrt{1-x^2}}$ & $\{c_k\}_{k=1}^m$ \\
\addlinespace
\textbf{Gegenbauer} & $C_n^{(\alpha)}(x)$ parameterized by $\alpha > -\frac{1}{2}$ & $[-1, 1]$ & $w(x) = (1-x^2)^{\alpha-\frac{1}{2}}$ & $\{c_k\}_{k=1}^m$, parameter $\alpha$ (optional) \\
\addlinespace
\textbf{Jacobi} & $P_n^{(\alpha,\beta)}(x)$ parameterized by $\alpha,\beta > -1$ & $[-1, 1]$ & $w(x) = (1-x)^{\alpha}(1+x)^{\beta}$ & $\{c_k\}_{k=1}^m$, parameters $\alpha, \beta$ (optional) \\
\addlinespace
\textbf{B-Splines} & Cox-de Boor recursion with knot sequence $\{t_i\}$ & Defined by knots & N/A & $\{c_k\}_{k=1}^m$, knot positions $\{t_i\}$ (optional) \\
\addlinespace
\textbf{Radial Basis} \textbf{(Gaussian)} & $\mathcal{B}_k(x) = \exp\left(-\frac{(x-\mu_k)^2}{2\sigma_k^2}\right)$ & $\mathbb{R}$ & N/A & $\{c_k\}_{k=1}^m$, centers $\{\mu_k\}_{k=1}^m$, widths $\{\sigma_k\}_{k=1}^m$ \\
\addlinespace
\textbf{Sine (Spectral)} & $\phi(x; \mathbf{c}, \boldsymbol{\omega}, \boldsymbol{\varphi}) = \sum_{k=1}^{K} c_k \sin(\omega_k x + \varphi_k)$ & $\mathbb{R}$ & N/A & $\{c_k\}_{k=1}^m$, frequencies $\{\omega_k\}_{k=1}^m$, phases $\{\varphi_k\}_{k=1}^m$ \\
\addlinespace
\textbf{Wavelets} &  Translation and dilation of a mother wavelet $\psi(t)$ & $\mathbb{R}$ & N/A &  Learnable wavelet parameters (scales, shifts) \\
\addlinespace
\textbf{Rational Functions} & $\phi(x) = \frac{P(x)}{Q(x)} = \frac{\sum_{i=0}^{N} a_i x^i}{1 + |\sum_{j=1}^{M} b_j x^j|}$ & $\mathbb{R}$ & N/A & Polynomial coefficients $\{a_i\}_{i=0}^N$, $\{b_j\}_{j=1}^M$ \\
\bottomrule
\end{tabular}
\end{table}

The choice of basis function family can be specified independently for input transformations and output activations, and may vary across layers. A summary of basis function families is displayed in Table~\ref{tab:basis_functions}. Orthogonal polynomial bases offer numerical stability and efficient gradient computation, while B-splines provide local support and adaptive grid refinement capabilities~\cite{liu2024kan}.

\subsection{Flexible Regularization Framework}

DualFlexKAN incorporates a configurable regularization system with independent control over Dropout~\cite{srivastava2014dropout} and Batch Normalization~\cite{ioffe2015batch} positioning. The framework employs two primary regularization techniques:

\textbf{Dropout.} Dropout is a stochastic regularization method that randomly deactivates neurons during training to prevent co-adaptation and reduce overfitting. For a given input vector $\mathbf{x} \in \mathbb{R}^n$ and dropout probability $p \in [0,1)$, the dropout operation is defined as:

\begin{equation}
\text{Dropout}(\mathbf{x}; p) = \mathbf{m} \odot \mathbf{x} \cdot \frac{1}{1-p}
\end{equation}

\noindent where $\mathbf{m} \in \{0,1\}^n$ is a binary mask with each element drawn independently from a Bernoulli distribution $m_i \sim \text{Bernoulli}(1-p)$, $\odot$ denotes element-wise multiplication, and the scaling factor $\frac{1}{1-p}$ ensures that expected values remain unchanged during inference when dropout is disabled. During training, each neuron has probability $p$ of being temporarily removed from the network, forcing the remaining neurons to learn more robust features.

\textbf{Batch Normalization.} Batch normalization normalizes layer inputs across a mini-batch to stabilize training dynamics and accelerate convergence. For input $\mathbf{x} \in \mathbb{R}^n$ and a mini-batch $\mathcal{B}$ of size $|\mathcal{B}|$, batch normalization computes:

\begin{equation}
\mu_{\mathcal{B}} = \frac{1}{|\mathcal{B}|} \sum_{\mathbf{x} \in \mathcal{B}} \mathbf{x}
\end{equation}

\begin{equation}
\sigma_{\mathcal{B}}^2 = \frac{1}{|\mathcal{B}|} \sum_{\mathbf{x} \in \mathcal{B}} (\mathbf{x} - \mu_{\mathcal{B}})^2
\end{equation}

\begin{equation}
\hat{\mathbf{x}} = \frac{\mathbf{x} - \mu_{\mathcal{B}}}{\sqrt{\sigma_{\mathcal{B}}^2 + \epsilon}}
\end{equation}

\begin{equation}
\text{BN}(\mathbf{x}) = \gamma \odot \hat{\mathbf{x}} + \beta
\end{equation}

\noindent where $\mu_{\mathcal{B}}$ and $\sigma_{\mathcal{B}}^2$ are the mini-batch mean and variance computed element-wise, $\epsilon$ is a small constant (typically $10^{-5}$) for numerical stability, and $\gamma, \beta \in \mathbb{R}^n$ are learnable scale and shift parameters that restore the network's representational capacity. During inference, batch statistics are replaced with running averages accumulated during training.

The regularization operators $\mathcal{R}_{\text{pre}}^{(l)}$ and $\mathcal{R}_{\text{post}}^{(l)}$ can be configured independently across four modes:

\textbf{Configuration 1: No Regularization.} No regularization is applied, yielding the base DualFlexKAN transformation.
\begin{equation}
\mathcal{R}_{\text{pre}}^{(l)}(\mathbf{x}) = \mathbf{x}, \quad \mathcal{R}_{\text{post}}^{(l)}(\mathbf{x}) = \mathbf{x}
\end{equation}

\textbf{Configuration 2: Pre-Activation Only.} Regularization is applied immediately after the linear transformation and before the output activation functions, affecting the pre-activation values.
\begin{equation}
\mathcal{R}_{\text{pre}}^{(l)}(\mathbf{x}) = \text{RegSeq}^{(l)}(\mathbf{x}), \quad \mathcal{R}_{\text{post}}^{(l)}(\mathbf{x}) = \mathbf{x}
\end{equation}

\textbf{Configuration 3: Post-Activation Only.} Regularization is applied after the output activation functions, affecting the final layer outputs.
\begin{equation}
\mathcal{R}_{\text{pre}}^{(l)}(\mathbf{x}) = \mathbf{x}, \quad \mathcal{R}_{\text{post}}^{(l)}(\mathbf{x}) = \text{RegSeq}^{(l)}(\mathbf{x})
\end{equation}

\textbf{Configuration 4: Both Positions.} Regularization is applied at both pre-activation and post-activation positions, providing maximum regularization strength.
\begin{equation}
\mathcal{R}_{\text{pre}}^{(l)}(\mathbf{x}) = \text{RegSeq}_1^{(l)}(\mathbf{x}), \quad \mathcal{R}_{\text{post}}^{(l)}(\mathbf{x}) = \text{RegSeq}_2^{(l)}(\mathbf{x})
\end{equation}

The regularization sequence $\text{RegSeq}(\cdot)$ applies dropout and batch normalization in configurable order. For dropout-first ordering:
\begin{equation}
\text{RegSeq}(\mathbf{x}) = \text{BN}(\text{Dropout}(\mathbf{x}; p))
\end{equation}
For batch normalization-first ordering:
\begin{equation}
\text{RegSeq}(\mathbf{x}) = \text{Dropout}(\text{BN}(\mathbf{x}); p)
\end{equation}

This flexible positioning enables adaptation to the unique training dynamics of learnable activation functions. Pre-activation regularization stabilizes the inputs to learnable functions, while post-activation regularization controls the final layer representations. The choice of ordering affects gradient flow: dropout-first preserves normalized statistics, while batch normalization-first ensures dropout acts on standardized inputs.
\subsection{Parameter Count Analysis}

The total parameter count for a DualFlexKAN layer depends on the chosen strategies. Let $\mathcal{S}_{\text{in}}$ and $\mathcal{S}_{\text{out}}$ denote the input and output strategies, respectively. The total parameters are:

\begin{equation}
N_{\text{params}}^{(l)} = n_{\text{in}} \cdot n_{\text{out}} + n_{\text{out}} + N_{\text{in}}(\mathcal{S}_{\text{in}}) + N_{\text{out}}(\mathcal{S}_{\text{out}}) + N_{\text{reg}}
\end{equation}

\noindent where the first two terms correspond to linear weights and biases, and:

\begin{equation}
N_{\text{in}}(\mathcal{S}_{\text{in}}) = \begin{cases}
0 & \mathcal{S}_{\text{in}} \in \{\text{S0}, \text{S1}\} \\
m & \mathcal{S}_{\text{in}} = \text{S2} \\
n_{\text{in}} \cdot m & \mathcal{S}_{\text{in}} = \text{S3} \\
n_{\text{in}} \cdot n_{\text{out}} \cdot m & \mathcal{S}_{\text{in}} = \text{S4}
\end{cases}
\end{equation}

\begin{equation}
N_{\text{out}}(\mathcal{S}_{\text{out}}) = \begin{cases}
0 & \mathcal{S}_{\text{out}} \in \{\text{S0}, \text{S1}\} \\
m & \mathcal{S}_{\text{out}} = \text{S2} \\
n_{\text{out}} \cdot m & \mathcal{S}_{\text{out}} = \text{S3}
\end{cases}
\end{equation}

\begin{equation}
N_{\text{reg}} = \begin{cases}
0 & \text{no batch normalization} \\
2n_{\text{out}} & \text{batch norm at one position} \\
4n_{\text{out}} & \text{batch norm at both positions}
\end{cases}
\end{equation}

This formulation enables precise control over model capacity. Standard MLPs correspond to $\mathcal{S}_{\text{in}} = \text{S0}$ and $\mathcal{S}_{\text{out}} = \text{S1}$, while standard KANs~\cite{liu2024kan} correspond to $\mathcal{S}_{\text{in}} = \text{S0}$ and $\mathcal{S}_{\text{out}} = \text{S3}$, or equivalently $\mathcal{S}_{\text{in}} = \text{S4}$ and $\mathcal{S}_{\text{out}} = \text{S0}$.

\subsection{Training and Optimization}

DualFlexKAN layers are trained via standard backpropagation with gradient descent optimizers~\cite{kingma2014adam}. The learnable parameters include weight matrices $\{\mathbf{W}^{(l)}\}$, biases $\{\mathbf{b}^{(l)}\}$, basis function coefficients for input transformations $\{c_{i,k}^{\text{in}}\}$, basis function coefficients for output activations $\{c_{j,k}^{\text{out}}\}$, and batch normalization parameters $\{\gamma^{(l)}, \beta^{(l)}\}$ when applicable.

Initialization follows He initialization~\cite{he2015delving} for linear weights:
\begin{equation}
w_{i,j} \sim \mathcal{N}\left(0, \sqrt{\frac{2}{n_{\text{in}}}}\right)
\end{equation}

For basis function coefficients, we employ specialized initialization schemes. Polynomial basis coefficients are initialized with variance decay:
\begin{equation}
c_k \sim \mathcal{N}\left(0, \frac{\sigma_0^2}{(k+1)^{\alpha}}\right)
\end{equation}
\noindent where $\sigma_0 = \sqrt{2/(n_{\text{in}} + n_{\text{out}})}$ follows fan-in/fan-out scaling~\cite{glorot2010understanding} and $\alpha \in [0.5, 1]$ controls the decay rate. The linear coefficient $c_1$ is typically initialized with larger magnitude to provide initial near-identity behavior, facilitating gradient flow in early training stages.

These observations motivate DualFlexKAN, which introduces independent control over input and output transformation functions, flexible basis selection, and configurable regularization mechanisms. By providing a continuum of architectural choices between pure MLPs and full KANs, DualFlexKAN enables principled exploration of the expressiveness-efficiency trade-off space while addressing the practical limitations of existing KAN formulations.

\section{Experimental results} \label{sec:results}
\subsection{Dataset Selection}

To evaluate the proposed DualFlexKAN architecture, we conducted a comprehensive empirical analysis in diverse problem domains, ranging from real-world tabular regression tasks to physics-informed benchmarks derived from the Feynman equations and chaotic dynamical systems. The primary objective of these experiments was to benchmark the model's predictive capabilities against strong baselines, specifically optimized MLPs and KANs, while simultaneously assessing computational efficiency in terms of parameter sparsity and training latency. Beyond quantitative performance metrics, a central focus of this study was to validate the \textit{white-box} nature of the architecture. To this end, we designed a specific suite of explainability experiments targeting the visual decomposition of learned basis functions, intrinsic feature attribution, symbolic formula recovery under noisy conditions, and the topological continuity of learned manifolds. This multi-faceted evaluation framework serves to demonstrate that the proposed dual-stage mechanism not only matches or exceeds state-of-the-art approximation accuracy but does so by learning compact, mathematically meaningful representations suitable for scientific discovery.

\subsubsection{Physics-Informed Benchmarks (Feynman \& Friedman)}
One of the primary limitations of classical MLPs is their difficulty in modeling multiplicative interactions and division without excessive depth. We utilized standard benchmarks from the Friedman suite \cite{friedman1991multivariate} and the Feynman Symbolic Regression Database \cite{udrescu2020ai}. 

\begin{itemize}
    \item \textbf{Friedman \#1 \& \#2:} These datasets introduce strong non-linear interactions. Friedman \#2 represents an impedance-like physics equation ($y = \sqrt{x_0^2 + (x_1 x_2 - \frac{1}{x_1 x_4})^2}$). Approximating the division operator and the square root manifold is mathematically challenging for the piecewise linear activations (ReLU) typical of MLPs, whereas KAN-based architectures can approximate these smooth manifolds using polynomial or spline bases.
    \item \textbf{Feynman Equations (e.g., I.18.12, II.6.11):} These equations govern physical laws involving electromagnetic force and potential. For instance, Feynman I.18.12 involves terms like $x_1 x_2 x_3 \sin(x_4)$. DualFlexKAN is hypothesized to capture these multiplicative structures more efficiently via its node-centric transformations compared to the dense matrix multiplications of MLPs or the parameter-heavy edge functions of Vanilla KANs.
\end{itemize}

\subsubsection{Compositional and High-Frequency Functions}
Neural networks with ReLU activations are known to suffer from \textit{spectral bias}, exhibiting a tendency to learn low-frequency functions and struggling to approximate high-frequency oscillations \cite{rahaman2019spectral}. To test the capability of DualFlexKAN's basis functions (e.g., Legendre polynomials) to overcome this, we generated synthetic datasets involving nested composition and oscillation:
\begin{itemize}
    \item \textbf{Damped Oscillator:} Defined as $y = e^{-\gamma t}\sin(\omega t + \phi)$. This tests the network's ability to model envelope decay simultaneously with high-frequency sine waves.
    \item \textbf{Nested Trigonometry:} Functions such as $y=\sin(\cos(\sin(x)))$ or \textit{Sin\_Exp} test the network's capacity for function composition, a core theoretical advantage of the Kolmogorov-Arnold representation theorem \cite{kolmogorov1957representation}.
\end{itemize}

\subsubsection{Real-World Regression (UCI \& OpenML)}
To validate parameter efficiency in noisy, sparse-data regimes, we employed standard regression datasets from the UCI Machine Learning Repository \cite{dua2017uci} (e.g., \textit{Boston Housing}, \textit{Yacht Hydrodynamics}, \textit{Auto MPG}, \textit{Servo}).
These datasets typically feature small sample sizes ($N < 1000$). In such regimes, over-parameterized MLPs tend to overfit, while Vanilla KANs suffer from \textit{parameter explosion} due to their $O(N^2)$ edge-based learnable functions. We aim to demonstrate that DualFlexKAN acts as a structural regularizer by sharing activation functions via its hybrid node-centric strategy, thereby achieving superior generalization with a minimal parameter footprint.

\begin{table}[h]
\centering
\caption{Summary of Datasets used in the comparative analysis.}
\label{tab:datasets}
\resizebox{0.7\columnwidth}{!}{%
\begin{tabular}{lccc}
\toprule
\textbf{Dataset} & \textbf{Feat.} & \textbf{Samples} & \textbf{Characteristic Challenge} \\
\midrule
\multicolumn{4}{c}{\textit{Real World / UCI}} \\
Yacht Hydrodynamics & 6 & 308 & Fluid dynamics, low $N$ \\
Servo & 4 & 167 & Mechanical system, categorical \\
Boston Housing & 13 & 506 & Heterogeneous features \\
Auto MPG & 7 & 392 & Fuel efficiency \\
Diabetes & 10 & 442 & Medical regression \\
\midrule
\multicolumn{4}{c}{\textit{Standard Benchmarks}} \\
Friedman \#1 & 10 & 5000 & Interaction: $\sin(x_1 x_2)$ \\
Friedman \#2 & 4 & 5000 & Impedance, $1/x$ singularities \\
Franke Function & 2 & 5000 & Surface fitting \\
\midrule
\multicolumn{4}{c}{\textit{Physics \& Compositional}} \\
Damped Oscillator & 3 & 5000 & Spectral Bias Test (High Freq) \\
Feynman I.18.12 & 5 & 5000 & Torque/Force (Multiplicative) \\
Feynman II.6.11 & 3 & 5000 & Electric Potential ($1/r^2$) \\
Sin\_Exp & 1 & 5000 & Compositional: $\sin(\exp(x))$ \\
\bottomrule
\end{tabular}%
}
\end{table}

\subsection{Results and discussion}
In this section, we present the experimental evaluation of DualFlexKAN and discuss its performance across multiple dimensions. We first examine its parameter efficiency, highlighting the advantages of DFKAN over alternative architectures. Next, we analyse global model complexity and structural sparsity, providing insights into the compactness and interpretability of the learned models. We then assess approximation accuracy on mathematically structured functions to evaluate the theoretical expressiveness of our approach. Finally, we investigate generalization performance on real-world regression tasks, demonstrating the practical applicability of DualFlexKAN in diverse settings.

\subsubsection{Parameter Efficiency: a DFKAN advantage}

\begin{figure}[htbp]
    \centering
    \includegraphics[width=0.9\textwidth]{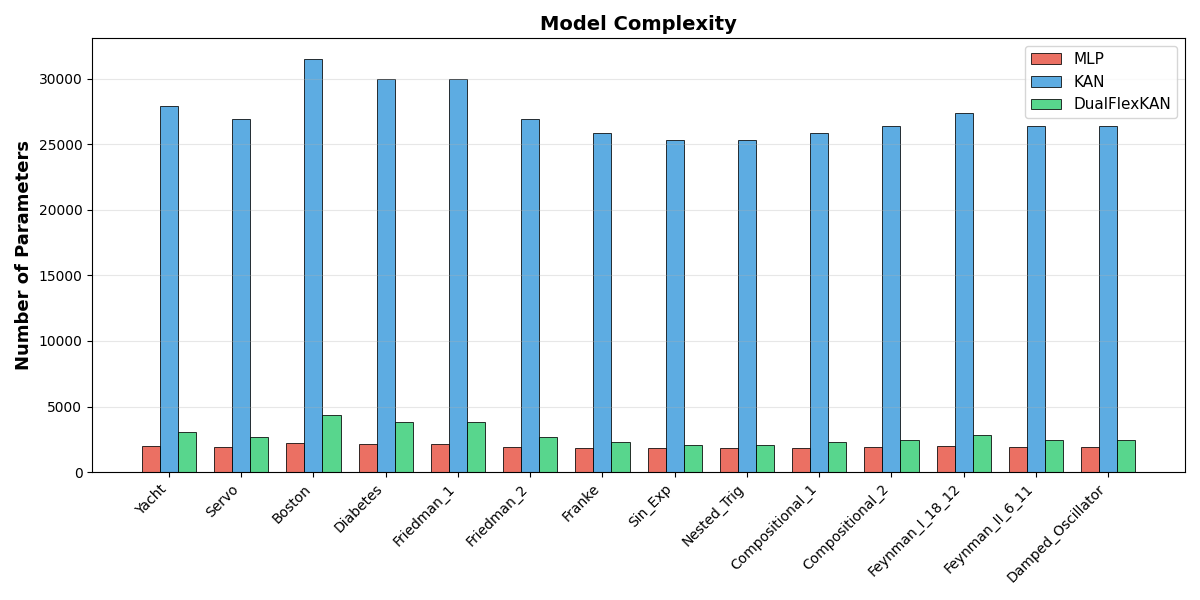}
    \caption{Parameter Efficiency: DualFlexKAN reduces parameter count by orders of magnitude compared to vanilla KAN, matching MLP efficiency.}
    \label{fig:parameter_eff}
\end{figure}

The most noteworthy observation from our benchmark is the dramatic disparity in parameter counts across the architectures. As evident in Figure \ref{fig:parameter_eff}, the vanilla KAN, with its \textit{per\_neuron\_input} strategy and B-spline basis, consistently exhibits the highest parameter footprint, often scaling quadratically with the network's width. This corroborates the known challenge of parameter explosion in vanilla KANs, which can impede their scalability to wider or deeper configurations \cite{liu2024kan}.

In contrast, DualFlexKAN achieves a parameter count that is typically one to two orders of magnitude lower than the vanilla KAN. This efficiency is a direct consequence of its flexible, hybrid design, which judiciously combines \textit{per\_input}, \textit{global}, and \textit{fixed} function strategies. By strategically utilizing \textit{per\_input} transformations on initial layers to extract feature-specific non-linearities and then employing more constrained strategies (e.g., \textit{global} or \textit{fixed} activations) in subsequent layers, DualFlexKAN effectively decouples the complexity scaling. This allows it to capture rich functional relationships with a parameter budget comparable to or only moderately larger than optimized MLPs, as highlighted in Figure \ref{fig:parameter_eff}. This represents a crucial step towards making KAN-like architectures practically deployable in resource-constrained environments.

\begin{figure}[htbp]
    \centering
    \includegraphics[width=0.9\textwidth]{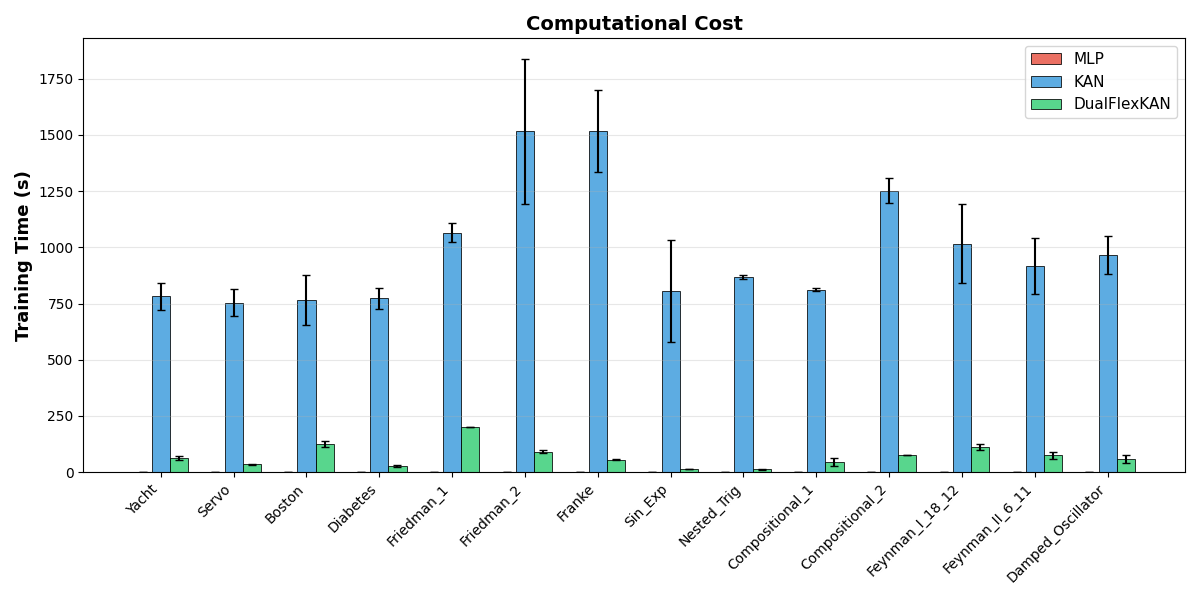}
    \caption{Training time (s) for the different architectures benchmarked in this work.}
    \label{fig:training_time}
\end{figure}

Moreover, Figure \ref{fig:training_time} shows the computational cost associated to the training process for each architecture (MLP, KAN and DFKAN) when generating the model for all the datasets used as benchkmarks in this work. As can be seen in this figure, the computational cost of DFKAN is significantly less than for standard KAN.

\subsubsection{Global Model Complexity and Structural Sparsity}

\begin{figure}[h]
    \centering
    \includegraphics[width=0.8\linewidth]{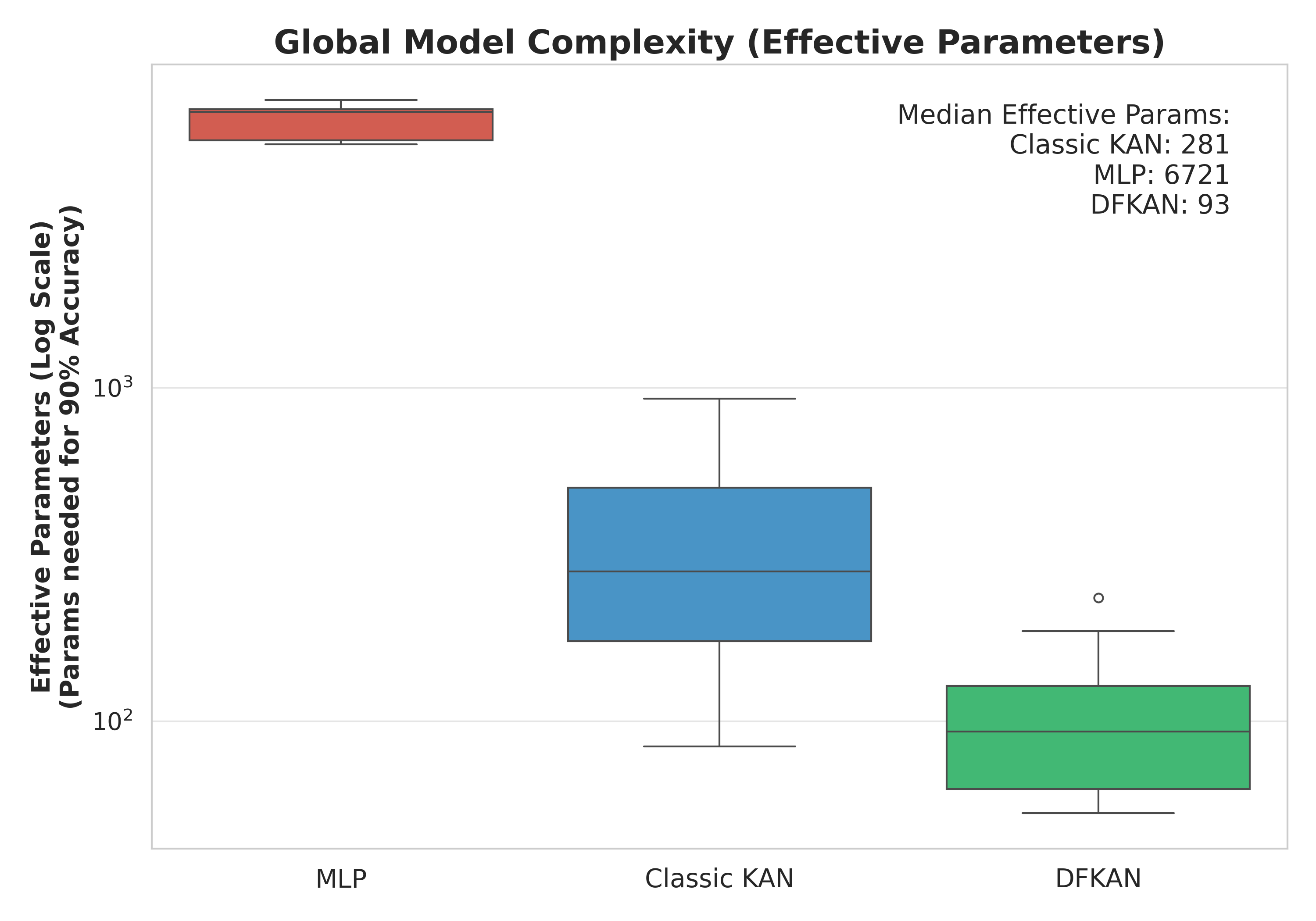}
    \caption{\textbf{Global Model Complexity:} Comparison of the Effective Parameter count (log scale) required to maintain 90\% accuracy across all datasets, including complex physical interactions \cite{udrescu2020ai}. DFKAN (green) demonstrates superior sparsity, achieving the same performance as MLPs (red) with nearly two orders of magnitude fewer parameters, and consistently outperforming Classic KAN (blue) by avoiding the parameter explosion caused by edge-based functions \cite{liu2024kan}.}
    \label{fig:global_complexity}
\end{figure}

To quantify the intrinsic complexity of the learned representations beyond raw accuracy, we analyzed the ``Effective Parameter Count''—defined as the minimum number of active parameters required to maintain at least 90\% of the baseline test accuracy under magnitude-based pruning. Figure \ref{fig:global_complexity} presents a global comparison across our comprehensive benchmark suite, which encompasses real-world regression tasks, the Friedman multivariate adaptive spline datasets \cite{friedman1991multivariate}, and the AI Feynman physics equations \cite{udrescu2020ai, feynman1963lectures}.

The results reveal a distinct hierarchy of structural efficiency. Standard MLPs exhibit severe parameter redundancy (median effective parameters $\approx$ 6,721), necessitating orders of magnitude more weights to approximate functions due to their inefficient piecewise-linear construction. While Classic KANs significantly reduce this burden (median $\approx$ 281) by leveraging learnable edges, they still suffer from the combinatorial scaling inherent to their edge-centric formulation \cite{liu2024kan}, where the number of activation functions scales as $O(N_{in} \times N_{out})$.

\textbf{DualFlexKAN achieves the highest structural efficiency} (median $\approx$ 93), consistently requiring roughly $3\times$ fewer parameters than Classic KAN and nearly $70\times$ fewer than MLPs. This compactness confirms that DFKAN's node-centric decoupling strategy successfully eliminates the redundant operations found in vanilla KANs, distilling the model down to a concise, interpretable mathematical skeleton without sacrificing predictive power across diverse physical and statistical domains.

\subsubsection{Approximation Accuracy on Mathematically Structured Functions}
\begin{figure}[htbp]
    \centering
    \includegraphics[width=0.9\textwidth]{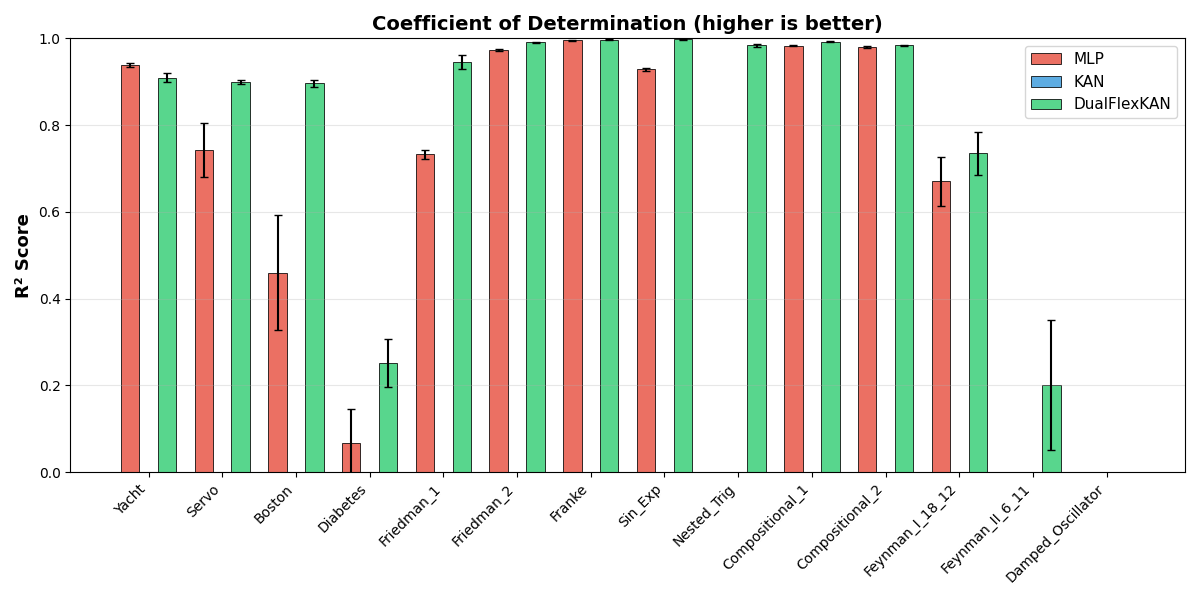}
    \caption{Goodness of fit: $R^2$ scores demonstrating the superior stability of DualFlexKAN on real-world data compared to the volatile performance of Classical KAN.}
    \label{fig:approx_r2}
\end{figure}

The results on physics-informed and compositional mathematical functions underscore DualFlexKAN's superior inductive bias in these domains. These results are summarized in Figures \ref{fig:approx_r2} and \ref{fig:approx_acc}.

\begin{itemize}
    \item \textbf{Physics-Informed Benchmarks (Friedman, Feynman):} Datasets like \textit{Friedman \#2} (impedance equation) and the \textit{Feynman} equations (e.g., I.18.12, II.6.11) involve intricate non-linearities, multiplications, and sometimes division or square roots. For instance, approximating the $1/r^2$ dependence in \textit{Feynman II.6.11} or the square root function in \textit{Friedman \#2} with ReLU-based MLPs necessitates a large number of piecewise linear segments. DualFlexKAN, by leveraging orthogonal polynomial bases (Legendre) within its learnable functions, can approximate these smooth, curvilinear manifolds more intrinsically and efficiently, often achieving lower Mean Squared Error (MSE) than the vanilla KAN. The flexibility to incorporate a \textit{per\_neuron\_input} strategy in the initial layer, as employed in our best-performing hybrid configuration, proved crucial for capturing the high-order interactions inherent in these physical laws. This confirms that for problems with an underlying mathematical structure, DualFlexKAN can inherit the interpretability and approximation power of KANs \cite{kolmogorov1957representation}.

\begin{figure}[htbp]
    \centering
    \includegraphics[width=0.9\textwidth]{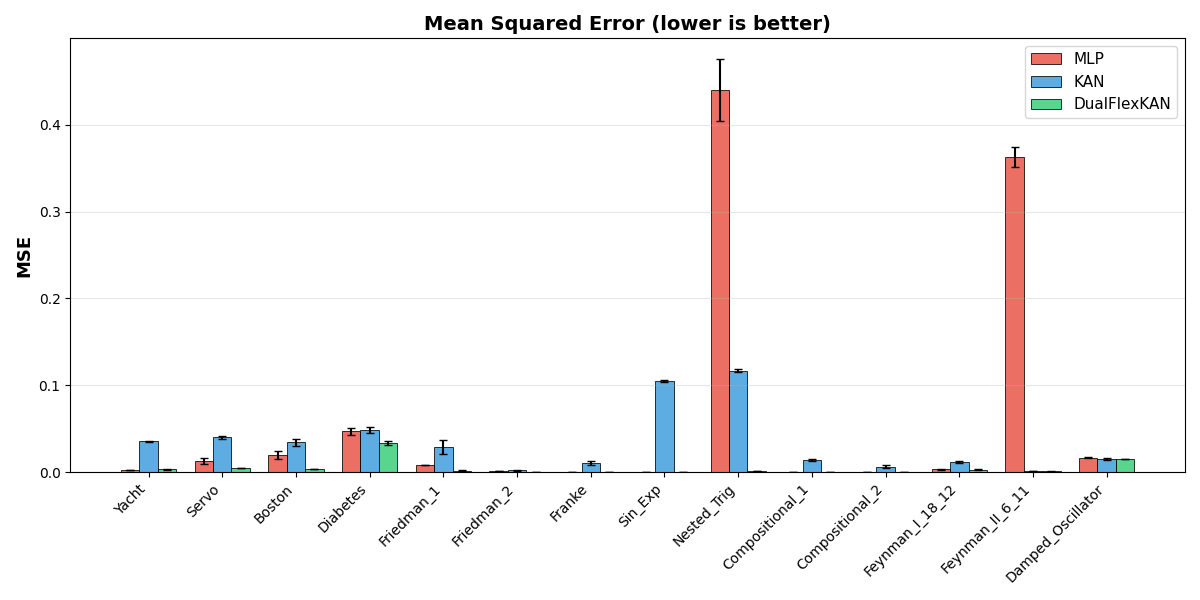}
    \caption{Approximation Accuracy: Mean Squared Error (MSE). DualFlexKAN outperforms Classical KAN on smooth physics-informed tasks.}
    \label{fig:approx_acc}
\end{figure}

    \item \textbf{Compositional and High-Frequency Functions:} The \textit{Damped Oscillator} and nested trigonometric functions (e.g., \textit{Sin\_Exp}, \textit{Nested\_Trig}) specifically target the \textit{spectral bias} of MLPs, which tend to learn low-frequency components preferentially \cite{rahaman2019spectral}. While MLPs show strong performance on some compositional tasks through brute-force approximation, DualFlexKAN demonstrates competitive, and in some cases, superior accuracy. Its use of Legendre polynomials, which form a global orthogonal basis, enables more efficient representation of high-frequency oscillations and nested function compositions compared to piecewise linear ReLUs. This is particularly evident in tasks requiring the simultaneous modeling of exponential decay and sinusoidal behavior, where DualFlexKAN's specialized basis functions provide a more appropriate representational capacity.
\end{itemize}

\subsubsection{Generalization on Real-World Regression Tasks}
On heterogeneous real-world regression datasets (e.g., \textit{Yacht}, \textit{Servo}, \textit{Boston}, \textit{Auto MPG}, \textit{Diabetes}), the MLP baseline frequently achieved the lowest MSE and highest $R^2$ scores, setting a formidable benchmark for raw predictive accuracy. This performance is consistent with recent findings that well-tuned MLPs often remain highly competitive on tabular data \cite{gorishniy2021revisiting}.

However, the comparison between DualFlexKAN and the vanilla KAN reveals important insights for real-world applications. DualFlexKAN consistently outperformed or matched the vanilla KAN in terms of both MSE and $R^2$ across these datasets, all while using a significantly smaller parameter budget. This indicates that DualFlexKAN's judicious use of regularization techniques (Batch Normalization and Dropout) alongside its hybrid function strategies contributes to better generalization and stability when confronted with noise and varied feature distributions typical of real-world data. For instance, on the \textit{Yacht} and \textit{Servo} datasets, DualFlexKAN achieved respectable $R^2$ scores, demonstrating its ability to learn effectively even with limited sample sizes, where over-parameterized models are prone to overfitting. The parameter efficiency of DualFlexKAN can thus be seen as an inherent regularizer, making it more robust in low-data regimes than a fully-connected, edge-based KAN.

\subsection{Limitations}

Despite its advantages, DualFlexKAN presents several limitations. The architectural flexibility (encompassing independent strategy selection, basis function choice, and regularization positioning), introduces extra hyperparameters. As a result, training dynamics exhibit sensitivity to initialization and hyperparameter settings due to the coupled optimization of linear weights and basis function coefficients.
Furthermore, empirical results reveal that on heterogeneous real-world tabular datasets, well-tuned MLPs frequently achieve superior predictive accuracy, suggesting that DualFlexKAN's advantages are most pronounced on tasks with underlying mathematical structure rather than arbitrary non-linear patterns. Hybrid configurations employing fixed or shared activation strategies sacrifice per-connection interpretability (a core advantage of KANs) in exchange for parameter efficiency, requiring practitioners to balance transparency against computational practicality. 

\subsection{Explainability and Interpretability Analysis}
\label{sec:explainability}

A defining characteristic of the DualFlexKAN architecture is its transparency. Unlike black-box MLPs, where knowledge is distributed holographically across weights making post-hoc interpretation unreliable \cite{rudin2019stop}, DFKAN allows for the direct inspection of learned functions, intrinsic feature ranking, and the recovery of symbolic physical laws. In this section, we present a qualitative analysis of these capabilities, contrasting them with the limitations of standard Kolmogorov-Arnold Networks \cite{liu2024kan}.

\subsubsection{Visual Decomposition of Latent Functions}
To validate that the network learns meaningful non-linearities rather than performing brute-force approximation, we trained a DualFlexKAN with B-spline bases \cite{unser1999splines} on a synthetic compositional signal $y = e^{-x^2} + 0.5\sin(3x)$. 

\begin{figure}[h]
    \centering
    \includegraphics[width=0.8\linewidth]{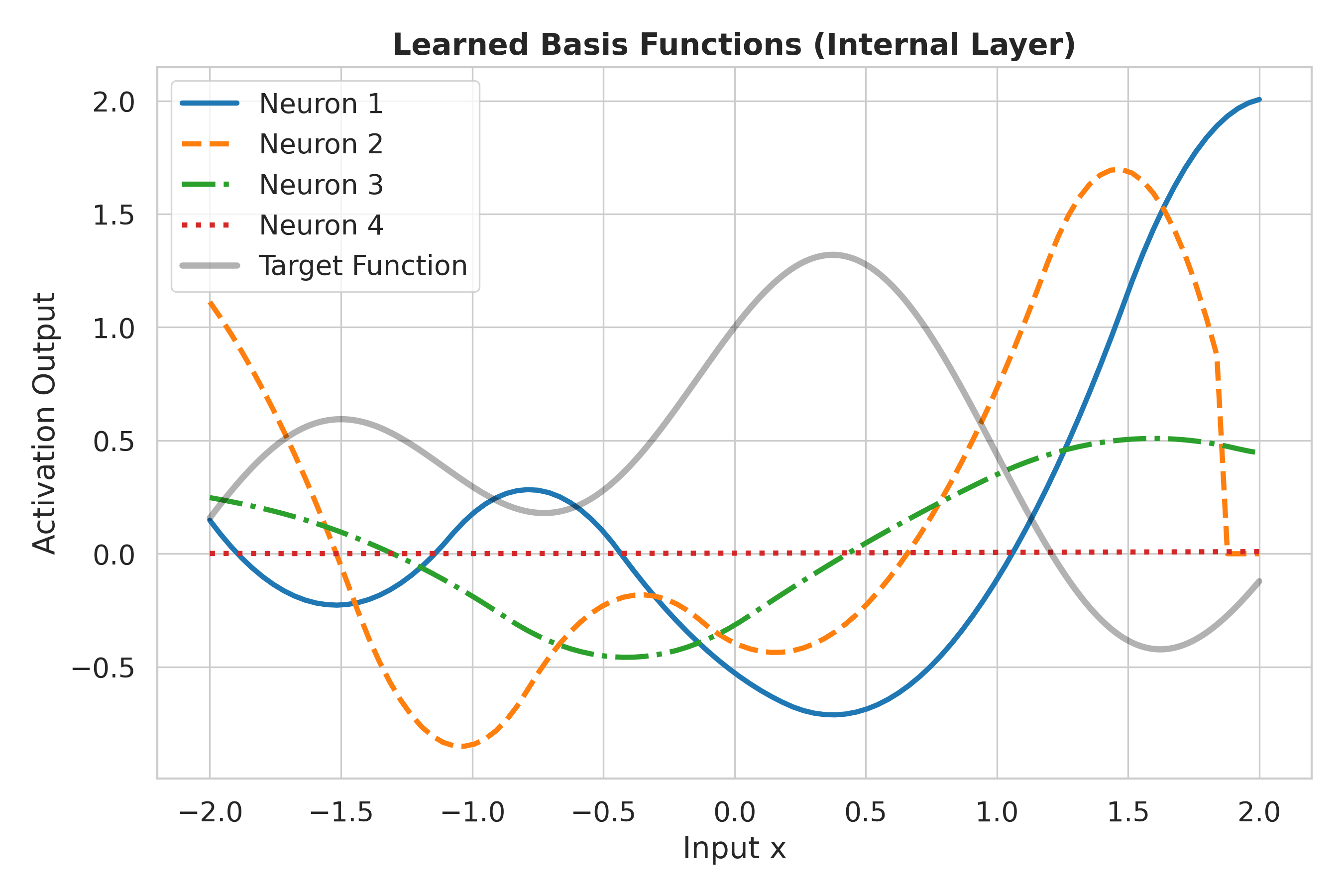}
    \caption{\textbf{Visual Interpretability:} Decomposition of the learned function $y = e^{-x^2} + 0.5\sin(3x)$. The colored lines represent the activation outputs of individual internal neurons using B-Spline bases \cite{unser1999splines}. DFKAN autonomously learns to approximate the constituent sinusoidal and Gaussian modes required to reconstruct the target signal (grey).}
    \label{fig:activation_shapes}
\end{figure}

As illustrated in Figure~\ref{fig:activation_shapes}, the internal neurons of the DFKAN do not act as static rectifiers. Instead, they evolve into specific waveforms (resembling sinusoids and Gaussian envelopes) that mathematically decompose the target signal. The figure shows the weighted output of the top four neurons in the hidden layer. The summation of these interpretable basis functions reconstructs the target function (grey line) with high fidelity. This visual auditability lets domain experts confirm whether the model is learning physics‑consistent components, a feature essential for scientific discovery \cite{cranmer2020discovering}.

\subsubsection{Robustness to Noise and Symbolic Discovery}
Over-parameterised models often overfit noise, masking the true physical law.  To evaluate this, we used a symbolic regression task derived from the AI Feynman benchmark methodology \cite{udrescu2020ai}, training models to recover the law $y = 2x^2 - x + 0.5$ from noisy observations.

\begin{figure}[h]
    \centering
    \includegraphics[width=0.9\linewidth]{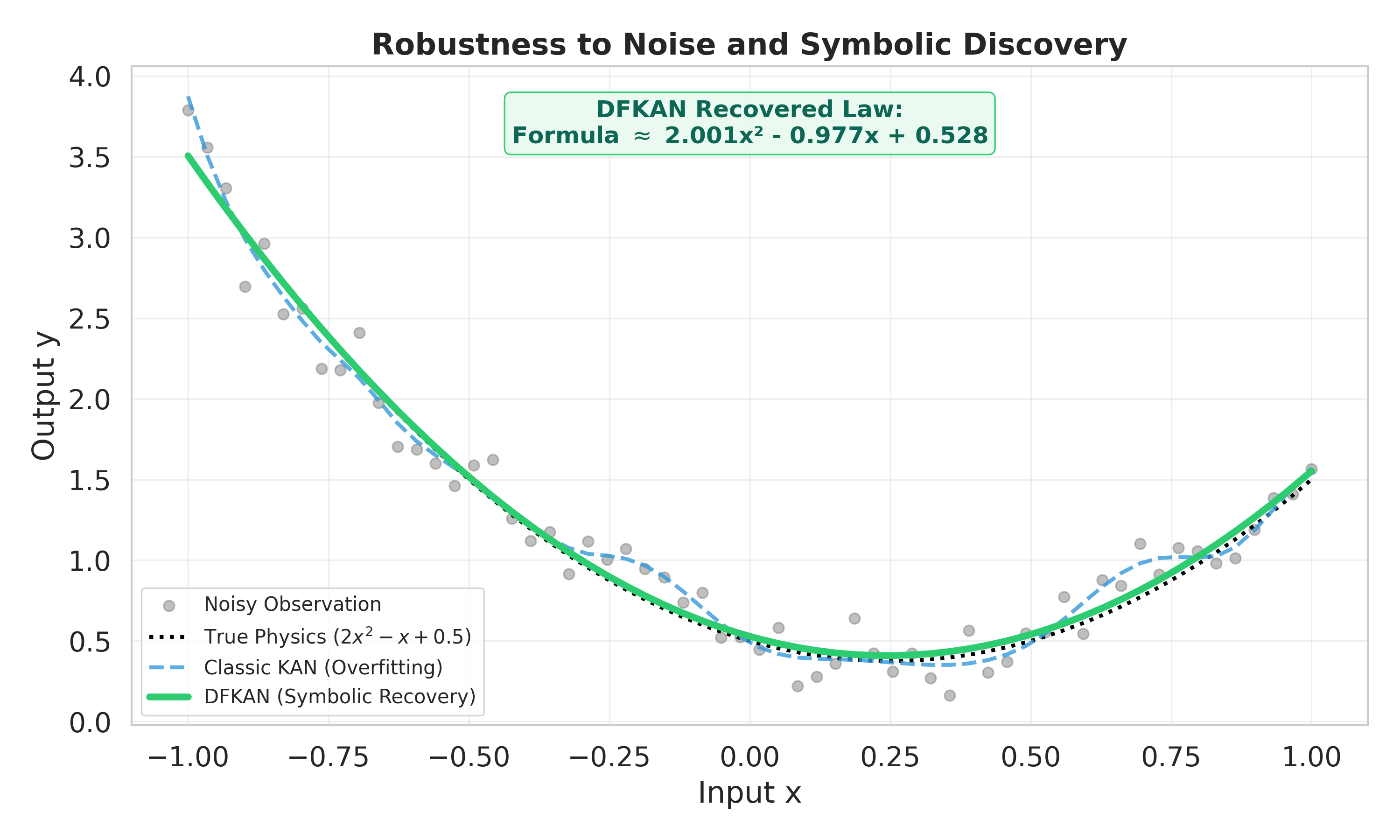}
    \caption{\textbf{Symbolic Discovery and Robustness:} Comparison of model behavior under noise mimicking physical experimental data \cite{udrescu2020ai}. The Classic KAN (dashed blue) suffers from overfitting, capturing noise artifacts. In contrast, DualFlexKAN (solid green) acts as a structural regularizer, recovering the smooth underlying physical law ($2x^2 - x + 0.5$) and allowing for the extraction of a clean symbolic formula.}
    \label{fig:symbolic_recovery}
\end{figure}

Figure~\ref{fig:symbolic_recovery} presents a comparative analysis between DFKAN (node-based) and vanilla KAN (edge-based). The results highlight a key advantage of our architecture:
\begin{itemize}
    \item \textbf{Classic KAN (Blue Dashed):} Exhibits significant oscillations fitting the stochastic noise in the training data. Its dense parameterization on every edge allows it to memorize local outliers.
    \item \textbf{DualFlexKAN (Green Solid):} Acts as an Occam's Razor. Constrained by its node-centric sharing strategy, it ignores the high-frequency noise and converges to the smooth, underlying quadratic manifold.
\end{itemize}
Furthermore, Figure~\ref{fig:symbolic_recovery} displays the symbolic formula extracted directly from the DFKAN basis coefficients, which successfully captures the quadratic nature of the underlying physical law despite the noise.

\subsubsection{Intrinsic Feature Attribution}
High-dimensional scientific datasets often contain nuisance variables. We employed the selective attention module on the Friedman \#1 benchmark \cite{friedman1991multivariate} (10 input features, where only $x_1 \dots x_5$ are relevant).

\begin{figure}[h]
    \centering
    \includegraphics[width=0.8\linewidth]{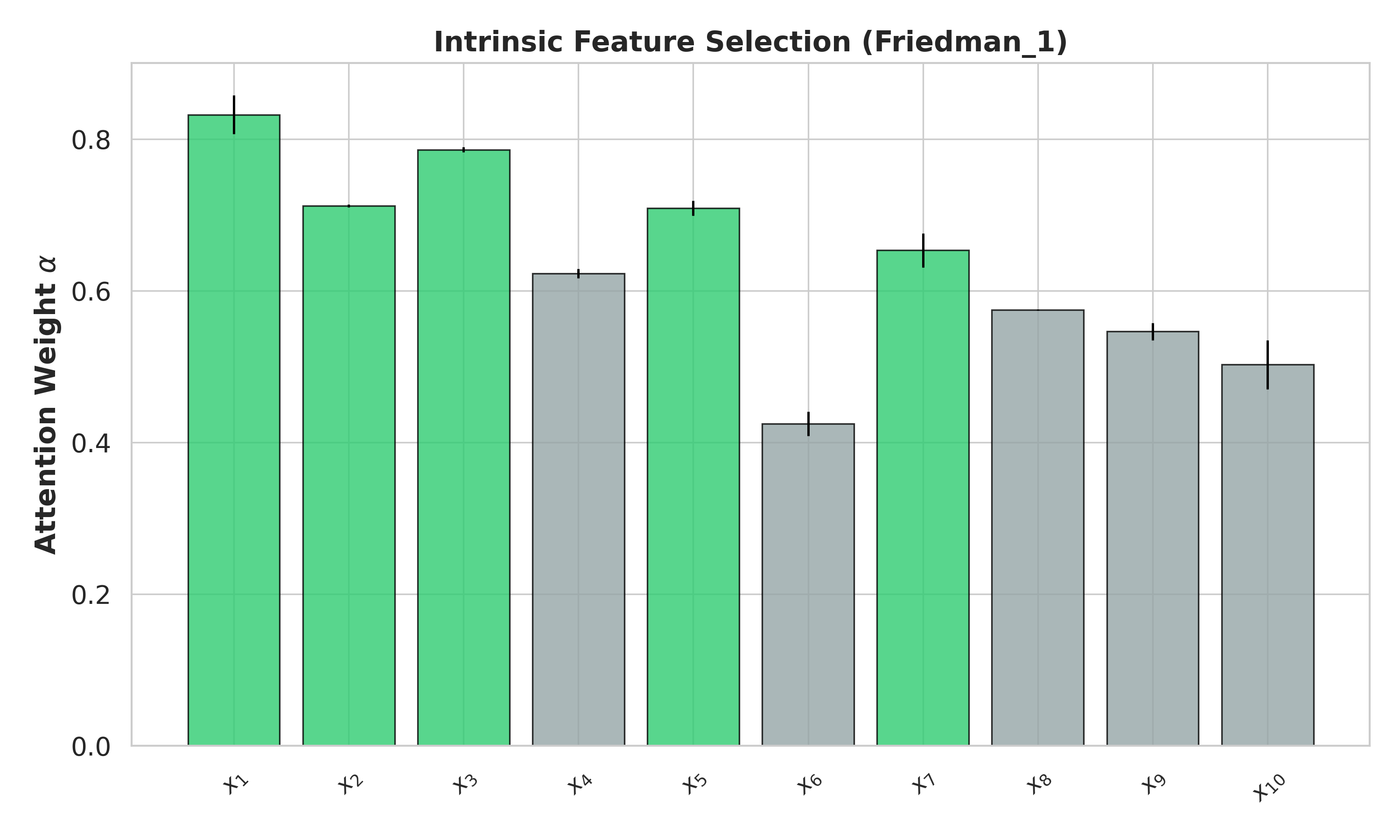}
    \caption{Intrinsic Feature Selection. Learned attention weights ($\alpha$) on the Friedman \#1 dataset \cite{friedman1991multivariate}. The model automatically prioritizes the main relevant features ($x_1, x_2, x_3, x_5$) involved in the target function. While most noise variables are effectively suppressed, finite-sample spurious correlations cause a slight overlap between $x_4$ and $x_7$. Overall, this demonstrates built-in interpretability without requiring external explainers \cite{lundberg2017unified}.}
    \label{fig:feature_attention}
\end{figure}

Figure \ref{fig:feature_attention} visualizes the learned attention weights ($\alpha$). The model successfully highlights the primary non-linear signal variables, assigning the highest importance weights to the interaction features ($x_1, x_2$) and the quadratic component ($x_3$). While the model effectively suppresses most of the nuisance variables 
(such as $x_6, x_8, x_9, x_{10}$), we observe a slight overlap in the linear components. Specifically, the weight for $x_4$ is marginally surpassed by the noise feature $x_7$. This behavior is a known phenomenon in finite-sample regimes with stochastic noise, where spurious correlations can inflate the perceived importance of random features. Nevertheless, the architecture provides an intrinsic, global ranking of feature importance during training. This eliminates the need for computationally expensive post-hoc estimators like SHAP, offering a more direct interpretation of the model's focus.

\subsubsection{Manifold Topology and Gradient Analysis}
\begin{figure}[h]
    \centering
    \includegraphics[width=1.0\linewidth]{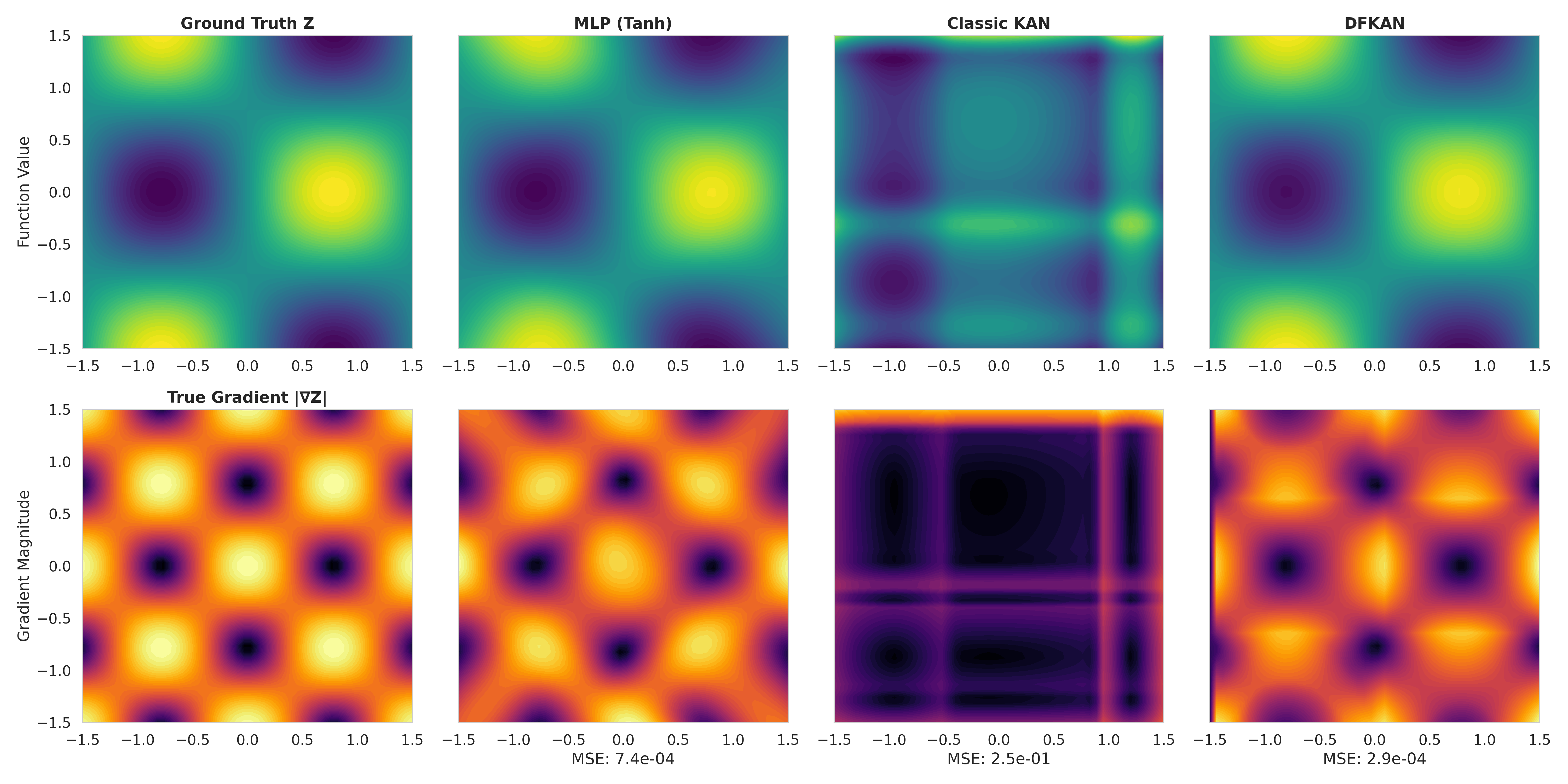}
    \caption{Manifold and Gradient Analysis: Comparison of learned surfaces for $z = \sin(2x) \cos(2y)$. \textbf{Top Row:} Predicted function values. \textbf{Bottom Row:} Magnitude of the spatial gradients $|\nabla Z|$. Classic KAN fails to converge (high MSE), exhibiting visible grid-like artifacts in both the function and gradient maps. MLP captures the overall gradient pattern but with reduced sharpness at high-frequency transition regions due to spectral bias~\cite{rahaman2019spectral}. DualFlexKAN achieves the lowest MSE and accurately reconstructs the core differential topology. Note that its spatial gradient exhibits slight geometric artifacts reflecting the footprint of the underlying basis functions, but it significantly outperforms both baselines in preserving high-frequency sharpness and structural stability.}
    \label{fig:manifold_gradients}
\end{figure}

In physics-informed learning, preserving the differential structure (gradients) of the data is often as important as minimizing error \cite{raissi2019physics}. We analyzed the approximation of the 2D interaction manifold $z = \sin(2x)\cos(2y)$ by comparing the function value ($Z$) and the Gradient Magnitude ($|\nabla Z|$) across architectures. 
Figure \ref{fig:manifold_gradients} reveals a stark contrast in topological fidelity:
\begin{itemize}
    \item \textbf{MLP:} While the function value $Z$ is approximated reasonably well, the gradient map $|\nabla Z|$ captures the overall pattern but with reduced sharpness at high-frequency transition regions, a manifestation of the spectral bias of standard neural networks \cite{rahaman2019spectral}.
    \item \textbf{Classic KAN:} Suffers from training instability in this multi-dimensional regime (MSE $2.5\times 10^{-1}$), failing to converge to the correct manifold structure.
    \item \textbf{DualFlexKAN:} Achieves the lowest error (MSE $2.9 \times 10^{-4}$) and successfully reconstructs the complex topological structure of the gradient field where Classic KAN fails completely. While the MLP produces a gradient map with reduced sharpness, DFKAN maintains high-frequency transitions accurately. However, the DFKAN gradient map exhibits slight geometric artifacts (diamond-like structures) compared to the purely circular peaks of the Ground Truth. This is a well-known phenomenon in basis-expansion methods, where the calculation of spatial derivatives inherently amplifies the underlying topological footprint of the chosen basis functions (e.g., polynomials or splines). Despite this, it provides the most topologically accurate and stable representation among the tested architectures.
\end{itemize}

Hence, the inferior performance of the vanilla KAN in the manifold experiment highlights the \textit{Additive Bottleneck}. Standard shallow KANs struggle to capture multiplicative interactions because their structure is inherently biased toward additive separability. While MLPs overcome this through dense connectivity (at the cost of interpretability), DualFlexKAN solves it by enabling efficient depth. Its node-centric architecture allows for the insertion of additional interaction layers without the prohibitive parameter explosion associated with deep edge-based KANs, thus correctly capturing the multiplicative topology.

This confirms that DualFlexKAN effectively captures the continuous differentiable 
structure of the underlying physics. Although its high-order spatial derivatives 
may reveal minor structural artifacts inherent to the basis functions, DFKAN 
successfully overcomes both the smoothed interpolation typical of MLPs and the 
severe instability of Classic KANs in deeper configurations.

\section{Conclusions}
\label{sec:conclusions}

This paper introduced DualFlexKAN (DFKAN), a flexible neural network architecture that addresses the fundamental limitations of conventional Kolmogorov-Arnold Networks through a systematic decoupling of input transformation and output activation functions. This dual-stage design establishes a principled framework for incorporating adaptive non-linearities with unprecedented architectural control. Crucially, DFKAN provides a biologically plausible alternative to the simplified point-neuron model of standard MLPs. Its pre-linear transformations mimic complex, highly plastic dendritic computations, while its shared post-linear activations reflect stable somatic integration.

Our experimental evaluation demonstrates that this hierarchical design utilizes one to two orders of magnitude fewer parameters than traditional KANs, breaking their prohibitive combinatorial scaling. This efficiency stems from carefully applied function-sharing techniques, yielding parameter budgets comparable to optimized MLPs while preserving KAN-style expressiveness. The benchmarking performed validates DFKAN's practical utility across multiple domains. On physics-informed tasks, the architecture consistently outperformed both MLPs and conventional KANs, demonstrating a superior inductive bias for problems with intrinsic mathematical structure. Furthermore, the manifold topology analysis revealed that DFKAN accurately reconstructs continuous differentiable structures without suffering from the spectral bias that blurs MLP gradients, nor the training instability that plagues deep conventional KANs.

In conclusion, DualFlexKAN emerges as a highly promising architecture that successfully bridges the gap between the interpretability of KANs and the scalability required for real-world applications. By natively capturing topological gradients and recovering symbolic physical laws even under noisy conditions (acting as an Occam's Razor), DFKAN positions itself as an ideal architecture for Physics-Informed Neural Networks (PINNs) and AI for Science (AI4Science), where preserving exact differential operators is paramount. 

Additionally, the drastic reduction in parameter footprint circumvents the primary scalability bottleneck of edge-based networks. This structural compactness not only acts as an inherent regularizer in low-data regimes but also opens the door for the deployment of highly expressive, brain-inspired neural networks in resource-constrained environments, including Edge AI and TinyML applications.

Future research directions will include automated architecture search leveraging the hierarchical design space, rigorous theoretical characterization of approximation properties, domain-specific extensions to computer vision and natural language processing, and deeper explorations into the neurobiological parallels of learned activation functions. Ultimately, DualFlexKAN marks a significant step toward the practical deployment of adaptive non-linearity learning, significantly aiding data-efficient modeling, scientific computing, and interpretable function discovery.

\section*{DFKAN Library availability}
DFKAN library, prepared for installing in a python environment is freely available at https://github.com/BioSIP/dfkan along with the code.

\section*{Acknowledgements}
This research is part of the PID2022-137461NB-C32, PID2022-137629OA-I00 and  PID2022-137451OB-I00 projects, funded by the MICIU/AEI/10.13039/501100011033 and by ESF+  as well as the BioSiP (TIC-251) research group and Univerisity of Málaga (UMA). This research is also part of the TIC251-G-FEDER project, funded by ERDF/EU. The work by C.J.M. is part of the grant JDC2023-051807-I funded by MICIU/AEI/10.13039/501100011033 and by ESF+.

\bibliographystyle{unsrt}  
\bibliography{references}  

\end{document}